\documentclass{article}

\usepackage[preprint]{neurips_2026}

\usepackage[utf8]{inputenc}
\usepackage[T1]{fontenc}
\usepackage{hyperref}
\usepackage{url}
\usepackage{booktabs}
\usepackage{amsfonts}
\usepackage{amsmath}
\usepackage{nicefrac}
\usepackage{microtype}
\usepackage{xcolor}
\usepackage{graphicx}

\title{Do Language Models Converge to Themselves?
Recursive Self-Refinement as Textual Relaxation}

\author{
Xuening Wu$^{1}$,
Qianya Xu$^{2}$,
Yanlan Kang$^{3}$,
Zeping Chen$^{4}$,
Yubin Liu$^{5}$,
Shenqin Yin$^{6,*}$\\[1.2ex]
$^{1}$Pfizer, Shanghai, China\\
$^{2}$University of California San Diego, La Jolla, CA, USA\\
$^{3}$Institute for Medical Philosophy and Future Artificial Intelligence, Baotou, Inner Mongolia, China\\
$^{4}$Tongji University, Shanghai, China\\
$^{5}$Shanghai Jiao Tong University, Shanghai, China\\
$^{6}$Institute of Humanities and Social Science Data, Fudan University, Shanghai, China\\[1ex]
$^{*}$Corresponding author: \texttt{ysq@fudan.edu.cn}
}

\begin{document}

\maketitle

\begin{abstract}
Large language models are increasingly used in recursive refinement workflows, where an initial draft is repeatedly revised by the same model. Despite the growing use of such workflows, their long-term dynamical behavior remains largely unexplored. Does repeated refinement continue to improve outputs indefinitely, or does it converge toward a stable textual form?
We study recursive self-refinement as a dynamical process, in which repeated LLM revision drives text toward a model-preferred soft fixed-point region. Using GPT-5.5, we generate refinement trajectories on 50 ICML 2025 abstracts over 10 iterations under both default-temperature and deterministic decoding, and additionally evaluate 15 ICML 2020 abstracts as a cross-year comparison. We characterize these trajectories using normalized edit distance, exact and approximate fixed-point statistics, word-count stability, exponential relaxation analysis, and external LLM-as-a-judge evaluation.
Across settings, refinement trajectories exhibit rapid saturation. Most edits occur within the first few iterations, after which trajectories enter a soft fixed-point region characterized by only small residual surface-level changes. Deterministic decoding reaches exact fixed points more rapidly and exhibits substantially smaller residual fluctuations than default-temperature decoding, while both settings achieve universal approximate convergence. The average edit magnitude $\Delta_t=d(V_t,V_{t+1})$ follows a consistent exponential relaxation pattern, suggesting a relaxation process toward a model-preferred textual equilibrium rather than open-ended optimization. External evaluation on sampled trajectories indicates that converged abstracts improve clarity, conciseness, and scientific style while preserving technical meaning.
These results support a dynamical-systems view of LLM self-refinement and motivate practical stopping criteria based on edit-magnitude saturation.
\end{abstract}

\section{Introduction}

Large language models (LLMs) are increasingly used not only to generate text, but also to revise their own outputs. In many practical workflows, a user or system provides an initial draft, asks the model to improve it, and then feeds the revised version back into the same model for further refinement. Such recursive self-refinement is common in scientific writing, coding, summarization, and agentic systems. Yet despite its growing use, its dynamical behavior remains poorly understood. Does repeated refinement continue to improve a text indefinitely, or does it relax toward a stable textual form?

Understanding recursive self-refinement is important for both practical and scientific reasons. Practically, users need to know how many refinement rounds are useful before additional iterations become redundant or potentially harmful. Scientifically, recursive self-refinement defines a discrete dynamical system over text: each model call maps one document version to another. This perspective raises natural questions about convergence, stability, attractors, residual fluctuations, and stopping criteria. However, most existing evaluations of LLM editing focus on single-step quality improvements rather than the dynamics induced by repeated self-application.

In this work, we study recursive self-refinement as \emph{textual relaxation}: a process in which repeated LLM revision drives an initial draft toward a model-preferred soft fixed-point region. Unlike open-ended optimization, textual relaxation predicts that edit magnitudes should rapidly decrease and eventually approach a small residual floor. This framing connects practical refinement workflows to dynamical-systems concepts such as fixed points, convergence rates, saturation, and stability under repeated self-application.

\begin{figure}[t]
    \centering
    \includegraphics[width=0.95\linewidth]{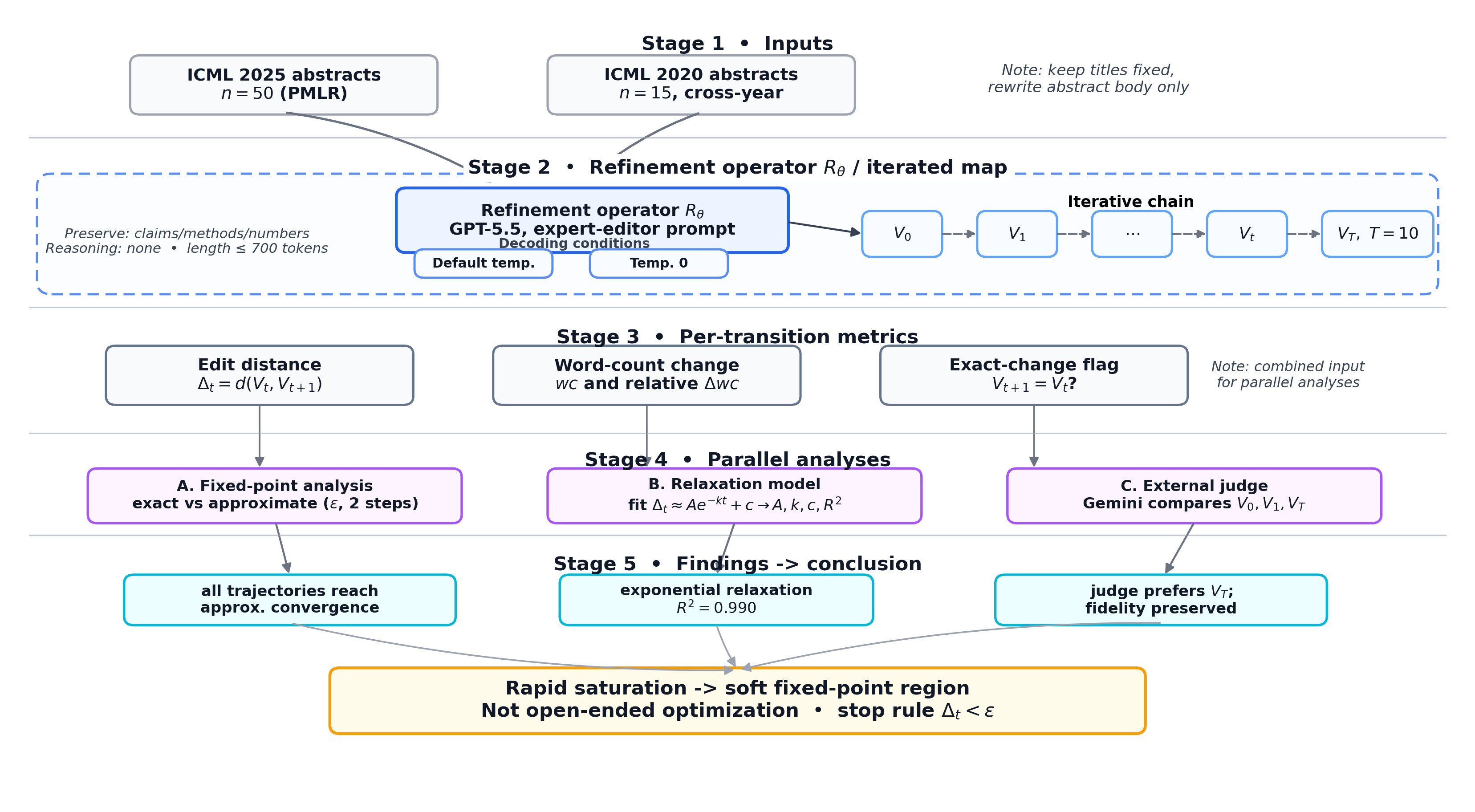}
    \caption{
    Overview of recursive self-refinement as a discrete dynamical system over text. Starting from an original abstract \(V_0\), the same refinement operator \(R_\theta\) is repeatedly applied to generate a trajectory \(V_0,V_1,\ldots,V_T\). We measure transition magnitudes \(\Delta_t=d(V_t,V_{t+1})\), analyze exact and approximate fixed points, and fit an exponential relaxation model \(\Delta_t \approx A e^{-kt}+c\). The residual term \(c\) captures a soft fixed-point floor, where further refinements produce only small surface-level fluctuations. An external Gemini judge compares \(V_0\), \(V_1\), and \(V_T\) to assess whether convergence corresponds to quality improvement rather than mere textual stagnation.
    }
    \label{fig:recursive_pipeline}
\end{figure}

We empirically examine this view in the domain of scientific abstract revision, where improvements in clarity and style must preserve technical meaning. Starting from ICML paper abstracts, we repeatedly ask GPT-5.5 to revise each abstract while preserving all claims, methods, results, and numerical values. This produces a trajectory of versions \(V_0,V_1,\ldots,V_T\), where \(V_0\) is the original abstract and \(V_{t+1}\) is the model's refinement of \(V_t\). We analyze these trajectories using normalized edit distance, word-count stability, exact fixed-point statistics, approximate convergence criteria, exponential relaxation fitting, and external LLM-as-a-judge evaluation.

A central challenge is that exact textual fixed points are too strict for natural language refinement. Even after a text is essentially stable, the model may continue to make small surface-level changes, such as punctuation, hyphenation, or minor phrasing adjustments. We therefore distinguish exact fixed points, where \(V_{t+1}=V_t\), from approximate fixed points, where consecutive versions differ only by a small edit distance and exhibit stable length. This distinction allows us to characterize recursive refinement as convergence to a soft fixed-point region rather than to a single canonical string.

Empirically, we find that recursive self-refinement converges rapidly. On 50 ICML 2025 abstracts refined for 10 iterations, GPT-5.5 reaches approximate convergence for all abstracts under both default-temperature and deterministic decoding. Under temperature \(0\), all abstracts also reach exact textual fixed points, with a mean exact convergence time of 2.6 iterations. The average transition magnitude \(\Delta_t=d(V_t,V_{t+1})\) decays sharply over time and is well described by an exponential relaxation curve, suggesting a relaxation process toward a model-preferred textual equilibrium rather than indefinite iterative optimization. Default-temperature decoding exhibits the same qualitative convergence behavior, but with larger residual fluctuations and later exact fixed points.

We further validate that convergence is not merely textual stagnation. An external Gemini judge evaluates sampled trajectories and consistently prefers the final refined abstracts over the original and first-refined versions, while assigning perfect technical-fidelity scores across versions. This suggests that recursive self-refinement improves clarity, conciseness, and scientific style without sacrificing technical meaning. A smaller cross-year experiment on ICML 2020 abstracts shows similar convergence behavior, indicating that the observed dynamics are not unique to contemporary LLM-era scientific writing.

Our contributions are threefold. First, we frame recursive LLM self-refinement as textual relaxation: a discrete dynamical process over text that approaches a model-preferred soft fixed-point region. Second, we provide empirical evidence that refinement trajectories rapidly saturate, with edit magnitudes following a regular exponential relaxation pattern. Third, we show that decoding temperature controls residual fluctuations around this region, and that edit-magnitude saturation motivates practical stopping criteria for recursive revision workflows. Together, these findings support a dynamical-systems view of LLM revision and suggest that iterative AI systems can be studied through the lens of convergence, stability, and textual equilibrium.

\section{Related Work}

\paragraph{Recursive self-refinement and reflection.}
Large language models are increasingly used in iterative refinement workflows, where an initial output is repeatedly revised through self-critique, reflection, or feedback. Self-Refine \citep{madaan2023selfrefine} introduced a framework in which a model generates feedback on its own outputs and iteratively revises them. Reflexion \citep{shinn2023reflexion} extended this idea to language agents that accumulate verbal feedback across attempts. Constitutional AI \citep{bai2022constitutional} similarly employs self-critique and revision guided by explicit principles. Collectively, these studies demonstrate that iterative refinement can improve output quality across diverse tasks. However, their primary focus is performance improvement after refinement, rather than the long-term dynamics induced by repeatedly applying the same model to its own outputs.

\paragraph{LLM revision and evaluation.}
Text revision systems are commonly evaluated using human judgments, semantic similarity measures, and automated metrics. More recently, strong language models have been shown to function as effective evaluators of generated text \citep{zheng2023judging, liu2023geval}. Such LLM-as-a-judge approaches are increasingly used to assess qualities such as clarity, helpfulness, style, and factual consistency. Scientific abstract refinement is a demanding setting because stylistic improvements must preserve technical meaning. We therefore combine trajectory-level convergence metrics with external LLM-based evaluation of clarity, conciseness, scientific style, and technical fidelity. This allows us to distinguish genuine refinement from mere textual stagnation.

\paragraph{Iterative reasoning and inference-time computation.}
Recent work has shown that repeated inference-time computation can improve language-model performance. Examples include Chain-of-Thought prompting \citep{wei2022chain}, Self-Consistency \citep{wang2023selfconsistency}, Tree-of-Thoughts \citep{yao2023tree}, and related inference-time scaling approaches. These methods explore, sample, evaluate, or revise intermediate reasoning paths to improve final-task accuracy. However, their primary objective is better task performance, not characterizing the trajectory produced by repeated self-application. In contrast, our work focuses on refinement trajectories themselves and asks whether repeated revision converges toward stable textual equilibria.

\paragraph{Dynamical systems, fixed points, and relaxation.}
Many iterative computational processes can be analyzed through the lens of dynamical systems, fixed points, convergence rates, attractors, and stability regions. Classical iterative algorithms are often characterized by how rapidly their residuals decay and whether they approach stable equilibria. Similar questions naturally arise in recursive LLM workflows: does repeated self-application converge, how quickly does the residual edit magnitude decay, and what form of equilibrium emerges? We refer to this behavior as \emph{textual relaxation}: a process in which repeated model revision drives text toward a model-preferred soft fixed-point region. Despite the growing adoption of recursive refinement in LLM systems, the relaxation dynamics of repeated revision remain largely unexplored.

\paragraph{Our perspective.}
In contrast to prior work, we treat recursive self-refinement itself as the primary object of study. Rather than asking only whether refinement improves final output quality, we investigate how refinement trajectories evolve under repeated application of the same refinement operator. We introduce exact and approximate fixed-point notions for text trajectories and provide empirical evidence that recursive self-refinement rapidly approaches a soft fixed-point region, with edit magnitudes following an exponential relaxation pattern. More broadly, our work shifts the focus from task-level improvement to the dynamical properties of recursive language-model behavior, including convergence, saturation, residual fluctuations, and practical stopping criteria.

\section{Method}

We study recursive self-refinement as a discrete dynamical process over text. Given an initial abstract \(V_0\), we repeatedly apply the same refinement operator \(R_\theta\), implemented by an LLM under a fixed prompt and decoding configuration:
\[
V_{t+1} = R_\theta(V_t), \quad t = 0,\ldots,T-1.
\]
This produces a refinement trajectory \(V_0,V_1,\ldots,V_T\). In our experiments, \(T=10\), and the prompt instructs the model to improve clarity, coherence, conciseness, and scientific writing quality while preserving all technical content, claims, methods, results, and numerical values. This setup lets us study the behavior induced by repeated self-application of the same revision operator.

\paragraph{Transition magnitude.}
To measure how much each refinement step changes the text, we compute the normalized Levenshtein edit distance between consecutive versions:
\[
\Delta_t =
\frac{\mathrm{EditDistance}(V_t,V_{t+1})}
{\max(|V_t|,|V_{t+1}|,1)}.
\]
We interpret \(\Delta_t\) as the transition magnitude or residual edit pressure at iteration \(t\). Smaller values indicate that consecutive versions are closer, while sustained small values indicate saturation of the refinement process. We also track word counts and relative word-count change to distinguish minor surface edits from substantive compression or expansion.

\paragraph{Exact and approximate fixed points.}
An exact fixed point occurs when the refinement operator leaves the text unchanged:
\[
V_{t+1}=V_t.
\]
However, exact equality is often too strict for natural language. Even after a text is essentially stable, a model may still alter punctuation, formatting, or local phrasing. We therefore define approximate convergence using both edit distance and length stability. A transition is approximately converged if
\[
\Delta_t < \epsilon_{\mathrm{edit}}
\]
and
\[
\frac{|\mathrm{wc}(V_{t+1})-\mathrm{wc}(V_t)|}
{\max(\mathrm{wc}(V_t),1)}
< \epsilon_{\mathrm{wc}}.
\]
We use \(\epsilon_{\mathrm{edit}}=0.02\) and \(\epsilon_{\mathrm{wc}}=0.03\). A trajectory is said to reach a soft fixed point at iteration \(t\) if this condition holds for two consecutive transitions starting at \(t\). This criterion captures convergence to a stable textual region rather than exact string equality, and provides a practical stopping rule for recursive refinement workflows.

\paragraph{Textual relaxation model.}
To quantify how edit magnitudes decay over repeated refinement, we fit an exponential relaxation model to the average transition magnitude:
\[
\Delta_t \approx A e^{-kt} + c.
\]
Here, \(A\) captures the initial edit pressure, \(k\) is the empirical convergence rate, and \(c\) represents a residual fluctuation floor near the soft fixed-point region. We use this model as a compact description of textual relaxation: a rapidly saturating process in which repeated revision approaches a model-preferred stable region. The model is not intended to imply that LLM refinement follows a literal physical law or performs gradient descent on an explicit objective.

\paragraph{External judge evaluation.}
Distance metrics capture textual stability but not whether refinement improves quality. We therefore perform an external LLM-as-a-judge evaluation on a subset of trajectories. A Gemini judge compares \(V_0\), \(V_1\), and \(V_T\) for each sampled abstract and rates clarity, conciseness, technical fidelity, scientific style, unnecessary change, and meaning preservation. This evaluation tests whether convergence corresponds to improved scientific writing rather than mere textual stagnation.

\section{Experiments}

\paragraph{Datasets.}
We evaluate recursive self-refinement on scientific abstracts from ICML proceedings. Our main dataset contains 50 abstracts from ICML 2025, collected from the PMLR proceedings. To test whether the observed dynamics are specific to recent LLM-era writing, we also collect a smaller cross-year comparison set of 15 abstracts from ICML 2020. Each example consists of a paper title and abstract. The title is provided to the model at every refinement step, while only the abstract is recursively revised.

\paragraph{Refinement model and prompt.}
Our main refinement model is GPT-5.5. For each abstract, we generate a trajectory of \(T=10\) refinement iterations. The model is prompted to act as an expert scientific editor preparing abstracts for a top-tier machine learning conference. The prompt asks the model to improve clarity, coherence, conciseness, and scientific writing quality while preserving all technical content, claims, methods, results, and numerical values. The model is instructed to return only the revised abstract.

\paragraph{Decoding settings.}
For ICML 2025, we run two decoding conditions: the API default-temperature setting and deterministic decoding with temperature \(0\). In both settings, we disable explicit reasoning by setting reasoning effort to none and cap the output length at 700 tokens. For the ICML 2020 comparison, we use the deterministic temperature \(0\) setting. All experiments use the same prompt and refinement horizon.

\paragraph{Evaluation metrics.}
For each transition \(V_t \rightarrow V_{t+1}\), we compute normalized Levenshtein edit distance, word count, relative word-count change, and whether the text changed exactly. For each trajectory, we report the first exact fixed point, the first approximate fixed point, the mean and maximum normalized edit distance, and the final-step normalized edit distance. Approximate convergence uses \(\epsilon_{\mathrm{edit}}=0.02\), \(\epsilon_{\mathrm{wc}}=0.03\), and requires two consecutive approximately converged transitions.

\paragraph{External judge evaluation.}
To assess whether textual relaxation corresponds to quality improvement, we perform an external LLM-as-a-judge evaluation on 10 sampled ICML 2025 trajectories from the temperature \(0\) condition. The judge is Gemini 3.5 Flash. For each sampled abstract, the judge compares \(V_0\), \(V_1\), and \(V_{10}\), scoring clarity, conciseness, technical fidelity, scientific style, unnecessary change, and meaning preservation on a 1--5 scale, and selecting the best version.

\section{Results}

\begin{table}[th]
\centering
\footnotesize
\begin{tabular}{lccccccc}
\toprule
Condition & \(n\) & Exact rate & Approx. rate & Mean \(t^*_{\mathrm{exact}}\) & Mean \(t^*_{\mathrm{approx}}\) & Final \(\Delta_t\) & Zero-final rate \\
\midrule
ICML 2025, default & 50 & 0.94 & 1.00 & 4.66 & 2.42 & 0.0055 & 0.56 \\
ICML 2025, temp. \(0\)  & 50 & 1.00 & 1.00 & 2.60 & 1.38 & 0.0003 & 0.92 \\
ICML 2020, temp. \(0\)  & 15 & 1.00 & 1.00 & 3.00 & 1.67 & 0.0010 & 0.80 \\
\bottomrule
\end{tabular}
\vspace{0.5em}
\caption{
Convergence summary for recursive refinement. Exact rate is the fraction of trajectories reaching exact textual fixed points. Approximate rate uses \(\epsilon_{\mathrm{edit}}=0.02\), \(\epsilon_{\mathrm{wc}}=0.03\), and two consecutive converged transitions. Final \(\Delta_t\) is the mean normalized edit distance at \(V_9 \rightarrow V_{10}\). Zero-final rate is the fraction of trajectories with no final-step textual change.
}
\label{tab:main_results}
\end{table}

\begin{figure}[t]
    \centering
    \includegraphics[width=0.75\linewidth]{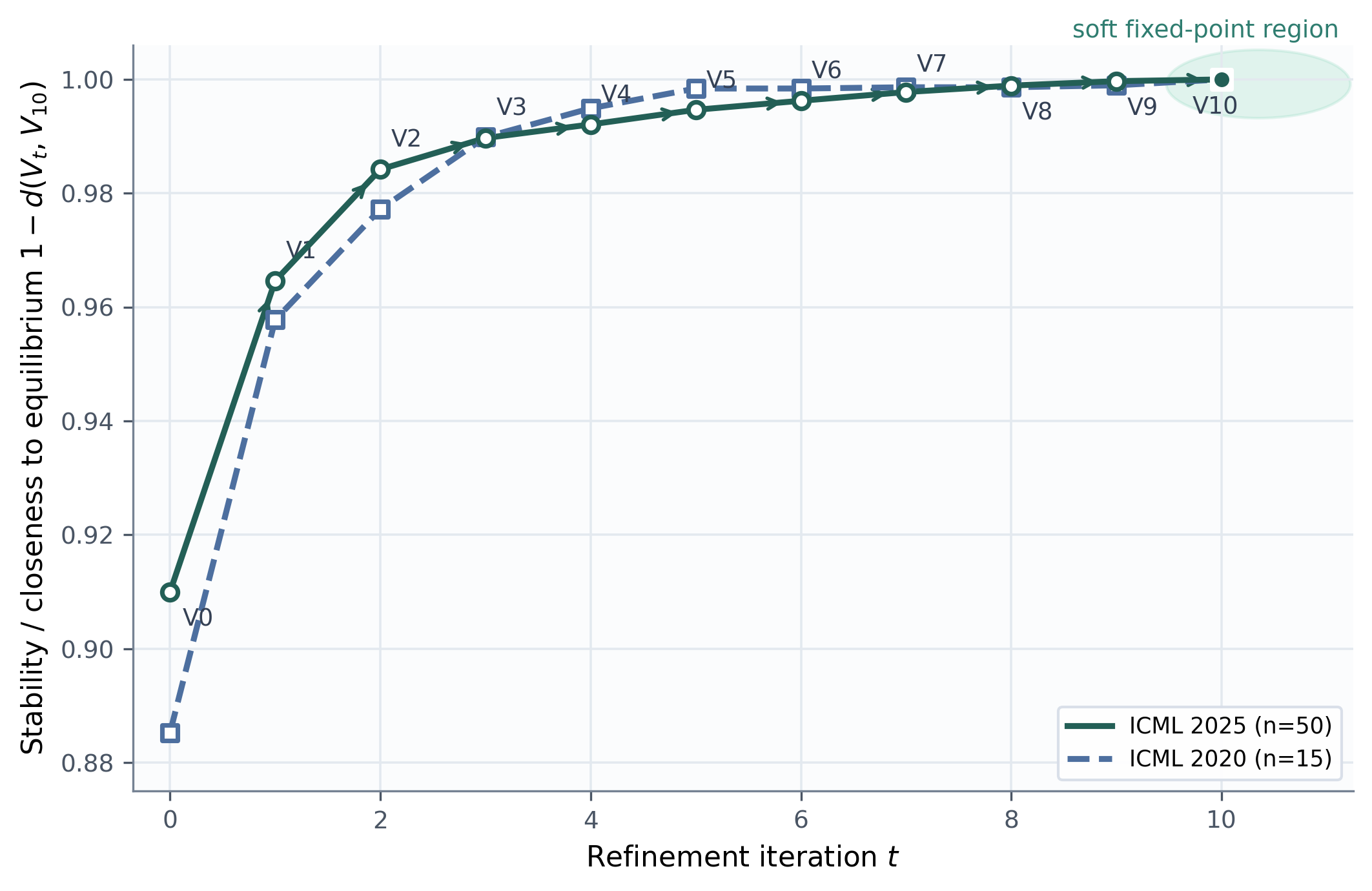}
    \caption{
Both ICML 2020 and ICML 2025 abstracts exhibit qualitatively similar convergence trajectories. Most movement occurs during the first few refinement iterations, after which trajectories rapidly approach a stable soft fixed-point region with only minor residual changes.
    }
    \label{fig:landscape}
\end{figure}

Figure~\ref{fig:landscape} visualizes the empirical convergence trajectory by projecting each refinement iteration according to its average closeness to the final version, \(1-d(V_t,V_{10})\). The trajectory rapidly enters a soft fixed-point region after the first few iterations, with only small residual movement from \(V_8\) to \(V_{10}\).

\paragraph{Recursive refinement converges rapidly.}
Table~\ref{tab:main_results} summarizes the main convergence results. On 50 ICML 2025 abstracts, GPT-5.5 reaches approximate convergence for all trajectories under both default-temperature and temperature \(0\) decoding. Exact fixed points are also common: 94\% of default-temperature trajectories and 100\% of temperature \(0\) trajectories reach exact textual fixed points within 10 iterations. Approximate convergence occurs earlier than exact convergence, with mean approximate convergence times of 2.42 and 1.38 transitions for default-temperature and temperature \(0\), respectively.

\begin{figure}[t]
    \centering
    \includegraphics[width=0.75\linewidth]{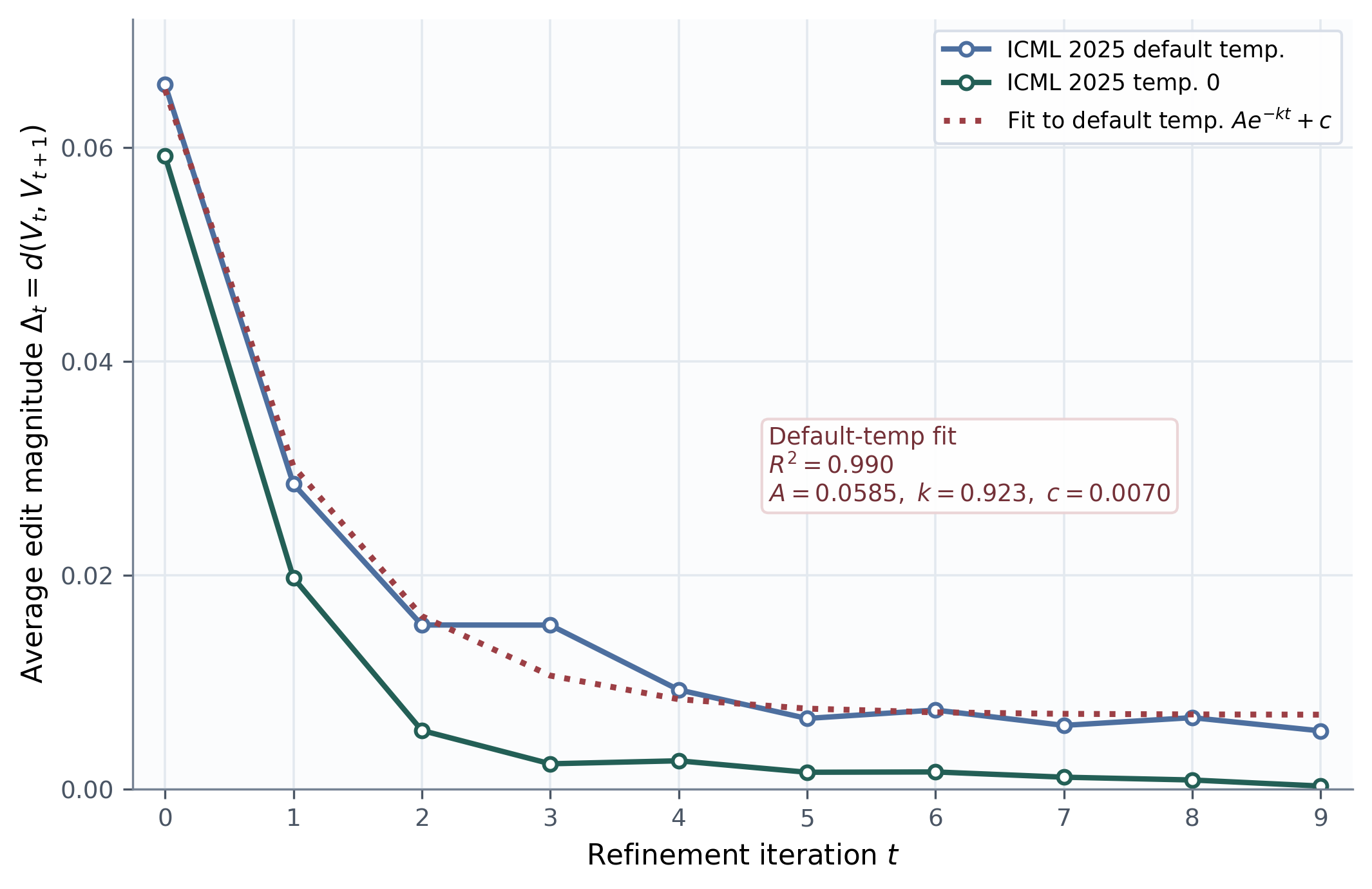}
    \caption{
    Exponential relaxation of recursive self-refinement. The average transition magnitude \(\Delta_t=d(V_t,V_{t+1})\) decays rapidly over early iterations and approaches a small residual floor. The default-temperature trajectory is well fit by \(\Delta_t \approx A e^{-kt}+c\) with \(R^2=0.990\), suggesting that recursive refinement behaves as a relaxation process toward a soft fixed-point region rather than an open-ended optimization procedure.
    }
    \label{fig:exponential}
\end{figure}

\paragraph{Temperature controls residual fluctuations.}
Figure~\ref{fig:exponential} shows that most editing occurs in the first few refinement steps. Under temperature \(0\), the mean normalized edit distance drops from 0.0592 at \(V_0 \rightarrow V_1\) to a final-step value of 0.0003 at \(V_9 \rightarrow V_{10}\). Default-temperature decoding exhibits the same qualitative decay but with larger residual fluctuations. Thus, deterministic decoding more often collapses to exact textual fixed points, while default-temperature decoding remains in a noisier soft fixed-point region.

\paragraph{Edit magnitudes follow a relaxation-like curve.}
The average transition magnitude is well fit by an exponential relaxation model,
\[
\Delta_t \approx A e^{-kt} + c.
\]
For the ICML 2025 default-temperature condition, the fitted curve achieves \(R^2=0.990\), with \(A=0.0585\), \(k=0.923\), and residual floor \(c=0.0070\). This supports the view that recursive refinement behaves like a rapid relaxation process toward a soft textual equilibrium rather than an open-ended optimization procedure. This saturation suggests a practical stopping criterion: recursive refinement can be terminated once \(\Delta_t\) remains below a small threshold for consecutive iterations.

\begin{table}[t]
\centering
\small
\begin{tabular}{lccccc}
\toprule
Version & Clarity & Conciseness & Technical fidelity & Scientific style & Unnecessary change \\
\midrule
\(V_0\)    & 4.0 & 3.9 & 5.0 & 3.5 & 1.0 \\
\(V_1\)    & 4.7 & 4.2 & 5.0 & 4.1 & 1.5 \\
\(V_{10}\) & 5.0 & 5.0 & 5.0 & 5.0 & 1.6 \\
\bottomrule
\end{tabular}
\caption{
Gemini judge evaluation on 10 sampled ICML 2025 trajectories under temperature \(0\). Scores are averaged across abstracts on a 1--5 scale. Gemini selects \(V_{10}\) as the best version for all 10 abstracts, with 100\% meaning preservation.
}
\label{tab:judge_eval}
\end{table}

\paragraph{External judging supports quality improvement.}
To verify that convergence is not merely textual stagnation, we evaluate 10 sampled trajectories using Gemini 3.5 Flash as an external judge. Gemini selects \(V_{10}\) as the best version for all 10 abstracts, while assigning perfect technical fidelity and meaning preservation across versions. Average clarity, conciseness, and scientific style all improve from \(V_0\) to \(V_{10}\), although unnecessary-change scores increase slightly, consistent with small residual stylistic churn near convergence.

\paragraph{Cross-year comparison.}
A smaller evaluation on 15 ICML 2020 abstracts under temperature \(0\) shows the same qualitative convergence pattern: all trajectories reach both approximate and exact fixed points. ICML 2020 abstracts have slightly larger initial edit distances than ICML 2025 abstracts, 0.0764 versus 0.0592, and slightly later exact convergence, with mean \(t^*=3.00\) versus 2.60. This suggests that the phenomenon is not unique to contemporary abstracts while hinting at possible temporal differences in scientific writing style.

\section{Discussion}

Our results suggest that recursive LLM refinement is better understood as textual relaxation than as open-ended optimization. Across abstracts and decoding settings, most edits occur in the first few iterations, after which trajectories enter a soft fixed-point region. This behavior is consistent with the view that repeated revision drives text toward a model-preferred region of scientific writing style. Once the text is close to this region, further refinements mostly produce small surface-level changes.

The observed exponential relaxation pattern is particularly noteworthy. Many iterative computational processes exhibit convergence, but the refinement trajectories studied here are well described by a simple exponential decay with a small residual floor. This pattern resembles an overdamped relaxation process: trajectories approach a stable region without evidence of persistent oscillation or divergence. We do not claim that LLM refinement follows a literal physical law, but the analogy provides a useful language for describing the observed dynamics: rapid early motion, saturation, and residual fluctuations near equilibrium.

This perspective also motivates practical stopping criteria. Although recursive refinement is sometimes described as an optimization-like procedure, our results suggest that its useful progress can be monitored directly through the residual edit magnitude \(\Delta_t=d(V_t,V_{t+1})\). A simple stopping rule is to terminate refinement once \(\Delta_t\) remains below a small threshold for consecutive iterations, optionally combined with word-count stability. This is analogous in spirit to convergence criteria used in iterative optimization, but does not require assuming that the model is explicitly performing gradient descent on a known objective.

The distinction between exact and approximate fixed points is important. Exact textual equality is a useful but overly strict criterion: a model may continue to adjust punctuation, hyphenation, or local phrasing even when the text is essentially stable. Approximate convergence better captures the practical stopping point of recursive refinement. In our experiments, approximate convergence occurs earlier and more consistently than exact convergence, suggesting that users may need only a small number of refinement rounds before additional iterations yield diminishing returns.

Temperature affects the nature of the fixed-point region. Deterministic decoding with temperature \(0\) quickly collapses to exact fixed points, while default-temperature decoding exhibits residual fluctuations around a similar soft fixed-point region. This supports the interpretation of temperature as controlling local variability near convergence rather than changing the overall direction of refinement.

More broadly, textual relaxation suggests that recursive LLM workflows may admit a richer stability analysis beyond surface edit distance. In a continuous representation space, one could study the local stability spectrum of refinement near a soft fixed-point region, characterizing which textual directions contract rapidly and which remain variable under repeated revision. Such an analysis could connect empirical edit-magnitude saturation to a more formal account of stability in recursive model use. It may also clarify whether refinement reduces trajectory-level variability as texts approach the fixed-point region.

These findings have practical implications for LLM-assisted writing workflows. Repeated refinement is not necessarily harmful, but it appears to have rapidly diminishing returns. For scientific abstracts, a few iterations are usually sufficient to reach a stable and improved version. Beyond that point, additional iterations may mostly introduce stylistic churn. This suggests that refinement systems should monitor convergence metrics and stop automatically once changes fall below a soft fixed-point threshold.
\section{Limitations}

Our study is limited in scope. We focus primarily on scientific abstracts from ICML proceedings and use GPT-5.5 as the main refinement model. The observed convergence behavior may differ for other genres, longer documents, more creative writing tasks, code refinement, multi-turn agentic workflows, or models with substantially different decoding behavior. Our cross-year comparison with ICML 2020 abstracts is smaller than the main ICML 2025 experiment and should be interpreted as a preliminary validation rather than a comprehensive temporal analysis.

Our convergence metrics are surface-based. Normalized edit distance and word-count stability are useful for measuring textual change, but they do not fully capture semantic equivalence, scientific correctness, or deeper rhetorical structure. We partially address this limitation with an external Gemini judge, but LLM-as-a-judge evaluation is itself imperfect and may reflect model-specific preferences. Human expert evaluation would provide a stronger assessment of whether refined abstracts preserve all technical claims while improving scientific communication.

The relaxation model is empirical. Although the exponential fit provides a compact description of the observed dynamics, it should not be interpreted as a physical law or as evidence that LLM refinement literally performs gradient descent on an explicit objective. The analogy to relaxation is intended to describe the observed shape of the refinement trajectory: rapid early change, saturation, and a small residual floor. Establishing a mechanistic theory would require stronger evidence, such as identifying an explicit objective, estimating local contraction properties, or analyzing the refinement operator in a continuous representation space.

Finally, our dynamical-systems analysis remains preliminary. We do not yet characterize the local stability spectrum of the refinement process in continuous representation spaces, nor do we directly analyze trajectory-level variability beyond surface edit distance. Such analyses are important directions for future work toward a fuller theory of textual relaxation.

\section{Conclusion}

We studied recursive self-refinement in large language models as \emph{textual relaxation}: a discrete dynamical process over text induced by repeated application of the same revision operator. Using GPT-5.5 refinement trajectories on ICML abstracts, we found that repeated self-refinement rapidly saturates. Most edits occur in the first few iterations, after which trajectories enter a soft fixed-point region with only small residual changes. Deterministic decoding reaches exact fixed points faster and with lower residual fluctuation than default-temperature decoding, while both settings achieve universal approximate convergence. The average edit magnitude is well described by an exponential relaxation curve, suggesting that recursive refinement behaves more like relaxation toward a model-preferred textual equilibrium than open-ended optimization.

These results suggest that approximate convergence metrics are useful for understanding and controlling iterative LLM workflows. In practical writing systems, monitoring edit distance and length stability can provide a simple stopping rule, preventing unnecessary refinement once the text has reached a stable region. More broadly, viewing LLM self-refinement through the lens of dynamical systems offers a compact framework for studying convergence, stability, residual variation, and soft fixed-point behavior in recursive model use. We hope this perspective motivates further work on the dynamics of iterative AI systems, including broader task domains, multi-model comparisons, and representation-space analyses of textual relaxation.

\bibliographystyle{plainnat}

\bibliography{references}

@inproceedings{wei2022chain,
  title={Chain-of-Thought Prompting Elicits Reasoning in Large Language Models},
  author={Wei, Jason and Wang, Xuezhi and Schuurmans, Dale and Bosma, Maarten and Xia, Fei and Chi, Ed H. and Le, Quoc V. and Zhou, Denny},
  booktitle={Advances in Neural Information Processing Systems},
  volume={35},
  pages={24824--24837},
  year={2022}
}

@inproceedings{madaan2023selfrefine,
  title={Self-Refine: Iterative Refinement with Self-Feedback},
  author={Madaan, Aman and Tandon, Niket and Clark, Peter and Yang, Yiming and others},
  booktitle={Advances in Neural Information Processing Systems},
  volume={36},
  year={2023}
}

@inproceedings{shinn2023reflexion,
  title={Reflexion: Language Agents with Verbal Reinforcement Learning},
  author={Shinn, Noah and Cassano, Federico and Gopinath, Ashwin and Narasimhan, Karthik R. and Yao, Shunyu},
  booktitle={Advances in Neural Information Processing Systems},
  volume={36},
  year={2023}
}

@article{bai2022constitutional,
  title={Constitutional AI: Harmlessness from AI Feedback},
  author={Bai, Yuntao and Kadavath, Saurav and Kundu, Sandipan and Askell, Amanda and others},
  journal={arXiv preprint arXiv:2212.08073},
  year={2022}
}

@inproceedings{zheng2023judging,
  title={Judging LLM-as-a-Judge with MT-Bench and Chatbot Arena},
  author={Zheng, Lianmin and Chiang, Wei-Lin and Sheng, Ying and Zhuang, Siyuan and Wu, Zhanghao and Zhuang, Yonghao and Lin, Zi and Li, Zhuohan and Li, Dacheng and Xing, Eric P. and others},
  booktitle={Advances in Neural Information Processing Systems},
  volume={36},
  year={2023}
}

@inproceedings{liu2023geval,
  title={G-Eval: NLG Evaluation using GPT-4 with Better Human Alignment},
  author={Liu, Yang and Iter, Dan and Xu, Yichong and Wang, Shuohang and Xu, Ruochen and Zhu, Chenguang},
  booktitle={Proceedings of the 2023 Conference on Empirical Methods in Natural Language Processing},
  pages={2511--2522},
  year={2023}
}

@inproceedings{wang2023selfconsistency,
  title={Self-Consistency Improves Chain of Thought Reasoning in Language Models},
  author={Wang, Xuezhi and Wei, Jason and Schuurmans, Dale and Le, Quoc V. and Chi, Ed H. and Narang, Sharan and Chowdhery, Aakanksha and Zhou, Denny},
  booktitle={International Conference on Learning Representations},
  year={2023}
}

@inproceedings{yao2023tree,
  title={Tree of Thoughts: Deliberate Problem Solving with Large Language Models},
  author={Yao, Shunyu and Yu, Dian and Zhao, Jeffrey and Shafran, Izhak and Narasimhan, Karthik and Cao, Yuan},
  booktitle={Advances in Neural Information Processing Systems},
  volume={36},
  year={2023}
}

\clearpage
\appendix

\section{Refinement Prompt}
\label{app:prompt}

The following prompt is used at every refinement iteration. The title is held fixed, while the abstract field is replaced by the current version \(V_t\).

\begin{list}{}{\leftmargin=1.5em\rightmargin=1.5em}
\item[]
\small
You are an expert scientific editor preparing abstracts for a top-tier machine learning conference.

\textbf{Task:} Revise the abstract to improve clarity, coherence, conciseness, and scientific writing quality.

\textbf{Requirements:}
\begin{enumerate}
\item Preserve all technical content, claims, methods, results, and numerical values.
\item Do not introduce new information.
\item Do not remove important information.
\item Do not change the meaning of any statement.
\item Prefer revisions only when they produce a genuine improvement.
\item If a sentence is already optimal, keep it unchanged.
\item Avoid unnecessary paraphrasing and synonym substitution.
\item Maintain a professional machine-learning research style.
\end{enumerate}

\textbf{Output:} Return ONLY the revised abstract.

\textbf{Title:} \{title\}

\textbf{Abstract:} \{abstract\}
\end{list}

\end{document}